\documentclass[lettersize,journal]{IEEEtran}
\usepackage{etoolbox}
\patchcmd{\IEEEtransactionjcmk}{\hfil}{}{}{} 
\usepackage{amsmath,amsfonts}

\usepackage[linesnumbered,ruled,vlined,boxed]{algorithm2e}
\SetAlgoLined
\DontPrintSemicolon
\usepackage{array}
\usepackage[caption=false,font=normalsize,labelfont=sf,textfont=sf]{subfig}
\usepackage{textcomp}
\usepackage{stfloats}
\usepackage{url}
\usepackage{hyperref}

\usepackage{verbatim}
\usepackage{graphicx}
\usepackage{cite}
\usepackage[capitalize]{cleveref}
\usepackage{booktabs}
\usepackage{multirow}
\usepackage{siunitx}
\usepackage{amsmath}
\usepackage{multirow}
\usepackage{makecell}

\usepackage{bm} 
\begin{document}

\title{DiCriTest: Testing Scenario Generation for Decision-Making Agents Considering Diversity and Criticality}

\author{Qitong Chu, Yufeng Yue, \IEEEmembership{Member,~IEEE}, Danya Yao, \IEEEmembership{Member,~IEEE}, Huaxin Pei
\thanks{Qitong Chu and Yufeng Yue are with the School of Automation, Beijing Institute of Technology, Beijing 100811, China.(email: chuqitong@bit.edu.cn, yueyufeng@bit.edu.cn)}
\thanks{Danya Yao and Huaxin Pei are with the Department of Automation, Tsinghua University, Beijing 100084, China. (email: 
yaody@tsinghua.edu.cn, phx17@tsinghua.org.cn).      (\textit{Corresponding author: Huaxin Pei}.)}
}



\maketitle

\begin{abstract}
The growing deployment of decision-making agents in dynamic environments increases the demand for safety verification. While critical testing scenario generation has emerged as an appealing verification methodology, effectively balancing diversity and criticality remains a key challenge for existing methods, particularly due to local optima entrapment in high-dimensional scenario spaces. To address this limitation, we propose a dual-space guided testing framework that coordinates scenario parameter space and agent behavior space, aiming to generate testing scenarios considering diversity and criticality. Specifically, in the scenario parameter space, a hierarchical representation framework combines dimensionality reduction and multi-dimensional subspace evaluation to efficiently localize diverse and critical subspaces. This guides dynamic coordination between two generation modes: local perturbation and global exploration, optimizing critical scenario quantity and diversity. Complementarily, in the agent behavior space, agent-environment interaction data are leveraged to quantify behavioral criticality/diversity and adaptively support generation mode switching, forming a closed feedback loop that continuously enhances scenario characterization and exploration within the parameter space. Experiments show our framework improves critical scenario generation by an average of 56.23\% and demonstrates greater diversity under novel parameter-behavior co-driven metrics when tested on five decision-making agents, outperforming state-of-the-art baselines.

\vspace{\baselineskip}

\end{abstract}

\section{Introduction}
\IEEEPARstart{D}{ecision-making} agents such as service robots and autonomous vehicles have rapidly advanced in capabilities through deep neural networks \cite{zhou2024indoor,nan2025safe,ou2023sim,hua2025multi,chai2022deep,cao2024deep,delgado2021fuzz}. In dynamic environments, however, their safety performance depends on real-time interaction reliability—a single incorrect decision may lead to catastrophic failures, threatening both agents stability and human safety\cite{feng2023dense}. This requirement necessitates comprehensive verification of agents across diverse testing scenarios to uncover critical scenarios before real-world deployment. Formally,  critical scenarios denote agent-environment interactions where either task failure occurs or maximum permitted execution attempts are exceeded\cite{10089194}.

The generation of diverse and critical scenarios has become an effective way for comprehensive testing of decision-making agents. Current state-of-the-art (SOTA) methods \cite{li2023generative, 10.1145/3597503.3639149, wang2023fuzzing} accelerate the generation of diverse critical scenarios through explicit diversity-aware mechanisms during exploration. Despite this advancement, their coverage remains severely limited in high-dimensional scenario parameter spaces, primarily due to the sparse distribution of critical scenarios. This limitation stems from two factors: non-uniform risk distribution geometrically confines critical scenarios to sparse subspaces, and the curse of dimensionality reduces random sampling effectiveness. To address these challenges, we must answer the pivotal question: How to devise targeted generation strategies to achieve comprehensive critical scenarios coverage in both high-dimensional scenario parameter space and agent behavior space?

\IEEEpubidadjcol

However, there are three critical barriers: (i) Critical scenarios localization in high-dimensional spaces remains intractable. The sparse distribution of critical scenarios in high-dimensional parameter spaces, coupled with the measure concentration phenomenon, leads to inefficient localization. While existing dimensionality reduction methods improve exploration speed, they fail to address complex parametric inter-dependencies. Our solution introduces a hierarchical representation framework that integrates three coupled metrics: spatial criticality for quantifying regional risk levels, test density for dynamically tracking under-explored regions, and neighborhood correlation for modeling local parametric dependencies, collectively enabling efficient identification of high-criticality regions. (ii) Inherent conflict between coverage completeness and search efficiency. Continuous exploration of new regions improves global coverage but reduces generation efficiency due to sparse critical scenarios distributions, while local over-exploration risks missing high-criticality regions. We propose a balancing strategy for testing scenarios exploration: In the scenario parameter space, our framework integrates the local perturbation mode and the global exploration mode to generate scenarios. The local perturbation mode achieves refined exploration through localized parameter space perturbations, while the global exploration mode performs cross-region exploration targeting diverse critical regions predicted by the hierarchical representation. In the agent behavior space, we devise a posteriori evaluation mechanism to compensate for scenario parameter space prediction deviations from actual agent behaviors. Agent-generated operational data enables accurate evaluation of scenario diversity, criticality, and generation strategies, directly driving scenario database updates and exploration mode selection. (iii) Lack of quantifiable diversity metrics. Current measures that rely on geometric distance or trajectory similarity ignore parameter-behavior interdependence. To address this, we propose parameter-behavior co-driven diversity metrics through joint quantification across the scenario parameter space and agent behavior space.

Based on the above analysis, this paper introduces DiCriTest, a decision-making agents testing framework that employs a dual-space guided testing mechanism to optimize the generation of diverse critical scenarios across both the scenario parameter space and agent behavior space. The framework implements a closed-loop workflow beginning with risk-aware initialization of scenario databases via uniform sampling, followed by density-driven critical scenarios generation through scenario parameter space guidance, and concluding with iterative refinement of the scenario database through coordinated spatial coverage and behavioral specificity metrics. 

Our main contributions can be summarized as follows: (i) We propose a scenario generation framework that accelerates testing scenarios exploration through scenario parameter space diversity and agent behavioral diversity, with guaranteed identification of critical scenarios. (ii) Our method achieves enhanced critical scenarios localization in high-dimensional parameter spaces through spatial criticality, test density and neighborhood correlation metrics, while adaptively optimizing the diversity-criticality trade-off via self-tuning parameters. (iii) We propose parameter-behavior co-driven diversity metrics, integrating three components: parameter space coverage, scenario distance and agent trajectory clustering divergence, which comprehensively characterize scenario diversity and enable efficient exploration guidance. (iv) Our method achieves a 56.23\% critical scenario generation rate improvement against SOTA baselines in five benchmarks, while demonstrating significant enhancements under our proposed parameter-behavior co-driven diversity metrics.

\section{Related Work}
\subsection{Testing of Decision-making Agents}
The evaluation of decision-making agents necessitates rigorous testing frameworks capable of exposing algorithmic weaknesses under diverse operational conditions. Previous research has largely focused on evaluating decision-making agents by modifying their inputs \cite{xie2019deephunter, tian2018deeptest, ma2018deepgauge}. For instance, DeepXplore \cite{pei2017deepxplore} employs a gradient-based search method to generate diverse and failure-inducing inputs, enabling the evaluation of deep learning models' decision-making capabilities. However, inputs generated by such models might be unrealistic and not reflect real-world condition. Additionally, some studies rely on historical data collected during testing \cite{ernst2019fast, lee2020adaptive, yamagata2020falsification}. Uesato et al. \cite{uesato2018rigorous}, for example, constructed a network model to learn patterns from past failure cases and predict new ones. However, historical data represents only previously observed failures and does not encompass the full range of possible situations. As a result, these methods often focus on known failure patterns while failing to identify new or unexpected issues, ultimately reducing the diversity and effectiveness of the testing process.

In contrast, generating testing scenarios provides a more realistic and versatile testing method. This approach enables the simulation of complex environments and facilitates the evaluation of decision-making performance under extreme or unforeseen conditions. For example, Menzel et al \cite{menzel2018scenarios} generates testing scenarios for autonomous vehicles based on expert experience. However, the above method rely on domain knowledge and have limited applicability. SCENORITA \cite{huai2023scenorita} generates testing scenarios for autonomous vehicles based on evolutionary algorithms. While this method effectively generates critical testing scenarios, it is limited to autonomous driving environments. Li et al. \cite{li2023generative} proposed generating testing scenarios based on diffusion models, which can be applied to a broader search space, offering stronger generalization ability and covering a wider range of testing scenarios.

While these methodologies have advanced the testing of decision-making agents, still have some limitations remain. These approaches either rely heavily on domain-specific expertise, limiting their generalization across different applications, or focus primarily on generating critical scenarios while neglecting the essential relationship between scenario diversity and criticality. To ensure comprehensive test coverage prioritizing specific rare high-impact scenarios, frameworks require unified optimization mechanisms that concurrently maintain scenarios diversity and criticality.

\subsection{Diversity and Criticality in Testing }

The exponential growth of decision-making agents' test space, driven by high-dimensional state transitions and complex environmental interactions, creates fundamental challenges in achieving both testing comprehensiveness and efficiency. Systematic generation of diverse yet critical testing scenarios enables efficient exploration of high-risk space within the vast test space, effectively balancing coverage breadth with fault detection effectiveness.

Recent methodologies for diverse critical scenario generation demonstrate distinct approaches across application domains and algorithmic frameworks. Existing research has introduced several methods to address scenario diversity. Domain-specific implementations, such as the AdvTest\cite{ma2024diversity} framework for StarCraft II multi-agent systems, preserve critical scenario diversity through dynamic constraint adaptation during critical states. MASTest \cite{ma2024enhancing}, focused on multi-agent collaboration and confrontation tasks, evaluates both individual and team diversity during testing.  Mazouni et al. \cite{mazouni2024testing} based on reinforcement learning, introduces a framework using Quality Diversity to address the issue of insufficient scenario diversity, particularly for failure modes. Bai et al. \cite{bai2024accurately} establish probabilistic benchmarks for high-risk scenario generation through a multi-stage learning framework that conducts probabilistic quantification of safety-critical events in high-dimensional state spaces. Concurrently, Mu et al. \cite{mu2024multi} achieve unified optimization of scenario diversity and criticality via a multi-agent reinforcement learning framework for autonomous driving, employing Hazard Arbitration Reward to filter non-responsible accidents and Scenarios Distinction Intrinsic Rewards to drive diverse exploration for efficient vulnerability discovery. Although these methods have contributed to improving scenario diversity in decision algorithm testing, they remain constrained by domain specificity and narrow decision-agent applicability.

\begin{figure*}[!t]
\centering
\includegraphics[width=18cm]{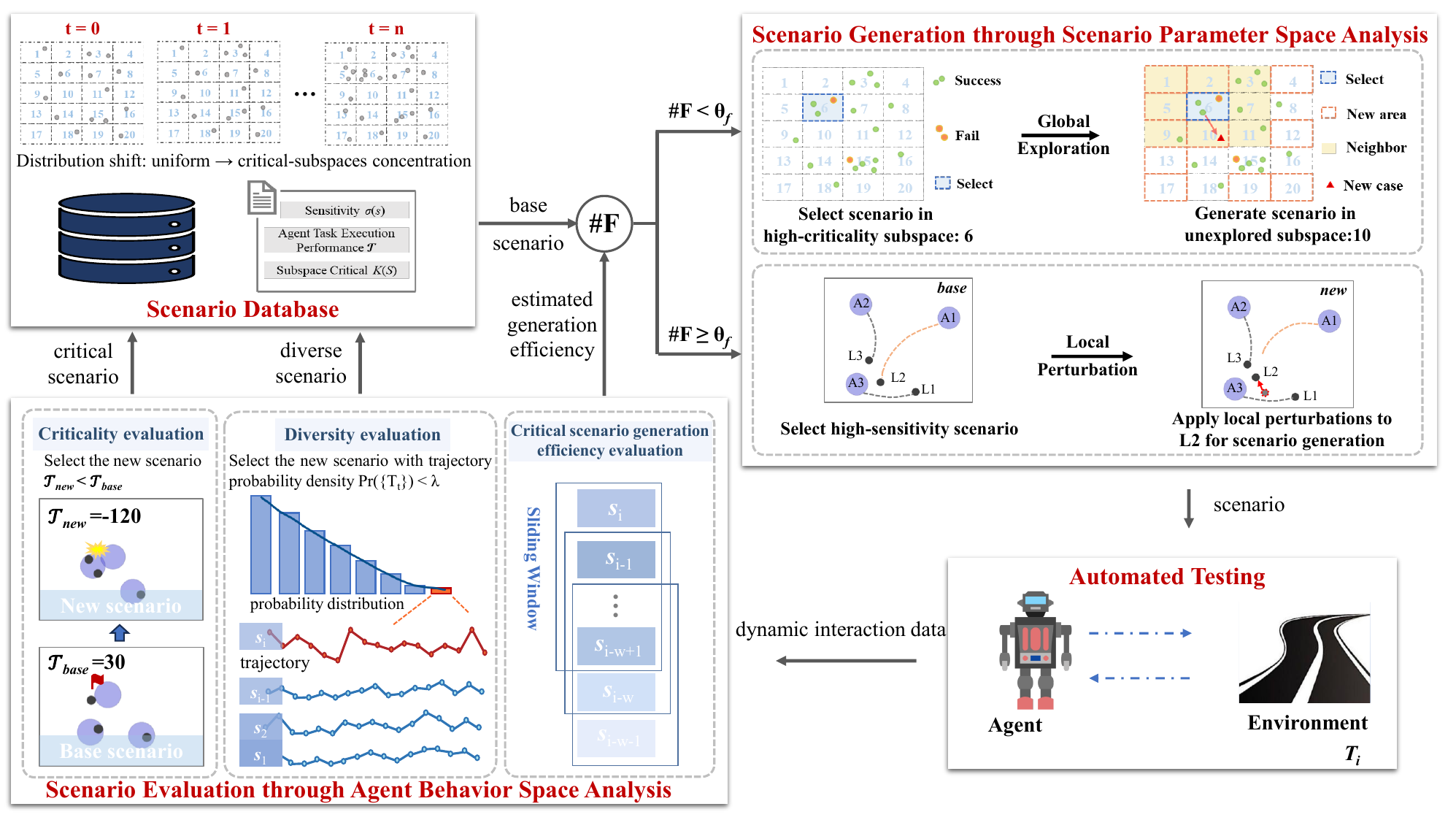}%
\caption{Workflow of our dual-space guided testing framework, namely DiCriTest. \textit{Scenario database} provides base scenarios and testing-related information for scenario generation. \textit{Scenario generation} combines parameter space analysis with database-provided base scenarios to explore critical and diverse testing scenarios. To address the issues of invalid resource consumption from repetitive local region exploration and insufficient coverage completeness in high-dimensional spaces, a scenario characterization framework is established to reduce exploration dimensions and predict diverse and critical sub-regions, while employing a combined approach of local perturbation mode and global exploration mode to proactively investigate uncovered high-risk regions. \textit{Automated testing} tests decision-making agents in generated scenarios. \textit{Scenario evaluation} analyzes multi-dimensional spatial behavioral data of decision-making agents from automated testing, implementing a methodology that integrates single-scenario and multi-cycle evaluations. These evaluations trigger scenario database updates and guide generation mode adjustments.}
\label{fig:workflow}
\end{figure*}

The growing complexity of decision-making agents has necessitated the development of cross-domain testing frameworks capable of generating diverse critical scenarios for heterogeneous algorithms and environmental conditions. MDPFuzz \cite{pang2022mdpfuzz}, a general testing framework, assesses scenario diversity by analyzing the spatial distribution of time series data generated through interaction between the decision strategy and the environment. Building on this foundation, CUREFuzz \cite{10.1145/3597503.3639149} uses a trained prediction network to assess diversity of critical scenarios by comparing the output of predicted and target networks, prioritizing high-diversity scenarios. SeqDivFuzz \cite{wang2023fuzzing} predicts scenario diversity through statistical analysis of historical agent trajectories and terminates the generation of low-diversity scenarios to optimize testing efficiency. These methods introduce metrics to evaluate scenario diversity; however, they primarily focus on the diversity of agent behavior space, overlooking the diversity of the scenario parameter space itself.

\section{Methodology}
We propose the DiCriTest framework, which coordinates the scenario parameter space with the agent behavior space to enhance diverse and critical testing scenario generation. The overall testing workflow is described in Section III-A, followed by a detailed representation method for the scenario parameter space in Section III-B. Finally, Section III-C presents the dual-space guided testing scenario generation strategy. 

\subsection{Approach Overview}

The DiCriTest framework establishes a closed-loop testing architecture with four integrated components: scenario database, scenario generation, automated testing, and scenario evaluation. The framework's core innovation lies in its joint coordination between scenario parameter space and agent behavior space, enabling efficient discovery of both diverse and safety-critical scenarios.

The scenario parameter space is formalized as a-dimensional vector space where each dimension corresponds to an environmental parameter governing agent operations. Systematic combinations of these parameters construct testing scenarios that mathematically define environmental features and task constraints, enabling precise characterization of multidimensional parameter interactions. Before automated testing, spatial potential evaluation identifies parameter subspaces with high criticality likelihood, optimizing scenario sampling efficiency through targeted generation.

After automated testing, the agent behavior space accumulates dynamic interaction data capturing decision-making patterns and performance metrics. Multi-dimensional evaluation of these behavioral responses quantifies scenario criticality and diversity, forming an evidence base for scenario database refinement and scenario generation adjustment. This dual-source optimization mechanism collaboratively integrates before-automated-testing parameter space analysis with after-automated-testing behavioral evaluation, driving closed-loop adaptive refinement of scenario generation strategies that progressively enhance both scenario diversity and criticality detection capability. \autoref{fig:workflow} presents the proposed framework, with the core components described as follows:

\subsubsection{Scenario Database Construction}
The scenario database provides base scenarios along with testing-related metrics per scenario to support scenario generation. Building upon this foundation, the scenario generation mechanism applies targeted modifications to these base scenarios, thereby avoiding the inefficiency of random exploration within the high-dimensional scenario parameter space. Initially, the database is constructed through uniform sampling of the scenario parameter space in the pre-testing stage. Throughout the testing process, it is dynamically maintained and updated to continuously improve both criticality and diversity of scenarios within the database.

\subsubsection{Scenario Generation through Scenario Parameter Space Analysis}
The scenario generation explores diverse and critical testing scenarios through analysis of the scenario parameter space, starting from base scenarios provided by the scenario database. Base scenarios are selected according to their high sensitivity where minor parametric perturbations induce significant variations in agent behavior patterns. In our approach, the generation mechanism primarily relies on local perturbations applied to base scenarios, and strategically leverages their pre-identified sensitivity characteristics to efficiently trigger critical scenarios. 

However, due to the sparse distribution of critical scenarios in the high-dimensional parameter space, scenarios generated through localized perturbations remain confined to narrow regions around base scenarios, limiting their spatial coverage. Such constrained exploration induces repetitive exploration within local regions, reducing the generation efficiency of critical scenarios. Consequently, local perturbation alone cannot meet the requirements for scenario diversity and criticality under limited testing resources. 

To address this, a balance between global exploration and local perturbation of the scenario parameter space is implemented. Specifically, the selection between local perturbation mode and global exploration mode depends on real-time monitoring of critical scenario generation efficiency within a recent time window. Local perturbation mode is activated when the critical scenario generation efficiency consistently exceeds a predefined threshold. This condition indicates sustained production potential in the current parameter region, prompting continued refined perturbations within this region. Global exploration mode is triggered when the generation efficiency falls below the threshold continuously. Guided by the spatial distribution patterns of historical parameters, unexplored high-potential regions are first identified, and cross-region exploration is subsequently conducted to expand coverage of critical testing scenarios. The implementation details of the scenario generation methodology are comprehensively described in Section III-B and Section III-C.

\subsubsection{Automated Testing}
The scenarios generated during the scenario generation phase are interacted with the decision-making agent for testing. Real-time feedback within the agent behavior space is recorded for scenarios evaluation.

\subsubsection{Scenario Evaluation through Agent Behavior Space Analysis}
Scenario parameter space analysis leverages historical parameter distributions to guide exploration directions for cross-region generation. However, relying solely on historical spatial parameter distributions proves to be insufficient to accurately predict the dynamic behavioral responses of agents. The resulting deviations in behavioral prediction lead to reduced localization accuracy, particularly in diverse and critical regions. To mitigate this deviation, an a posteriori-driven analysis of the agent behavior space is introduced for scenario evaluation. This facilitates updates to the scenario database and enables online optimization of generation mode selection. 

Specifically, real-time updates to the scenario database are driven by quantifying a scenario criticality score based on agents' task execution feedback within the current single-scenario instance. However, accurately assessing two key metrics, diversity and generation efficiency of critical scenarios, is a challenge using only current data from a single scenario. Diversity evaluation requires quantifying statistical distributions across historical scenarios, while generation efficiency requires time-series monitoring of sustained generation capability. Consequently, a multi-cycle scenario evaluator is implemented, integrating two core components: (i) efficiency tracking for dynamically adjusting exploration modes based on historical critical scenario generation rates, and (ii) diversity-aware regulation for maintaining long-term spatial diversity in the scenario database. Details of scenario evaluation methodology are provided in Section III-C.

\subsection{Scenario Parameter Space Representation}

The sparse distribution of critical scenarios in high-dimensional parameter spaces, coupled with insufficient characterization information, results in computational complexity and efficiency bottlenecks during direct exploration. To address this challenge, we propose a hierarchical representation framework to efficiently localize diverse and critical scenarios, thereby establishing foundational references for subsequent scenario generation.

Our framework jointly represents the scenario parameter space through a space abstraction layer and a subspace evaluation layer. The space abstraction layer reduces dimensionality by mapping high-dimensional spaces to low-dimensional subspaces. This transforms the exploration paradigm from scenario-level to subspace-level analysis, mitigating exploration complexity. Building on these subspaces, the subspace evaluation layer extracts multi-dimensional representation information—specifically test density, spatial criticality, and neighborhood correlation—to drive the localization of two target subspace types: diversity subspaces and critical subspaces. Diversity subspaces are identified using test density, while critical subspaces are localized through spatial criticality and neighborhood correlation. The detailed representation model is introduced as follows.

\subsubsection{Space Abstraction Layer}

The inherently high-dimensional and vast scenario parameter space for decision-making agents renders exhaustive exploration of critical scenarios computationally infeasible. To address this, we propose a space abstraction method that decomposes the high-dimensional space into lower-dimensional subspaces, enabling targeted and efficient scenario exploration.

For an $N$-dimensional scenario parameter space $\mathcal{S} = \prod_{i=1}^N [a_i, b_i] \subset \mathbb{R}^N$, we construct low-dimensional subspaces through hypercube tessellation:
\begin{align}
p_i^{(j)} &= a_i + \frac{j}{m_i}(b_i - a_i), \quad \forall j \in \{0,1,\dots,m_i\} \label{eq:partition}, \\
S_{\mathbf{k}} &= \prod_{i=1}^N \left[ p_i^{(k_i)}, p_i^{(k_i + 1)} \right), \quad k_i \in \{0,1,\dots,m_i-1\} \label{eq:subspace},
\end{align}
where the dimension index $i \in \{1,\dots,N\}$ identifies parameter axes, with $m_i \in \mathbb{Z}^+$ specifying the number of uniform partitions along the $i^{th}$ dimension. Each division point $p_i^{(j)}$ along dimension $i$ is computed via linear interpolation between bounds $a_i$ and $b_i$. The subspace index vector $\mathbf{k} = (k_1,\dots,k_N)$ encodes positional coordinates in the partitioned grid, where $k_i$ selects the $k_i^{th}$ interval $[p_i^{(k_i)}, p_i^{(k_i+1)})$ along dimension $i$. 

\begin{figure}[!t]
\centering
\subfloat[]{\includegraphics[width=1.5in]{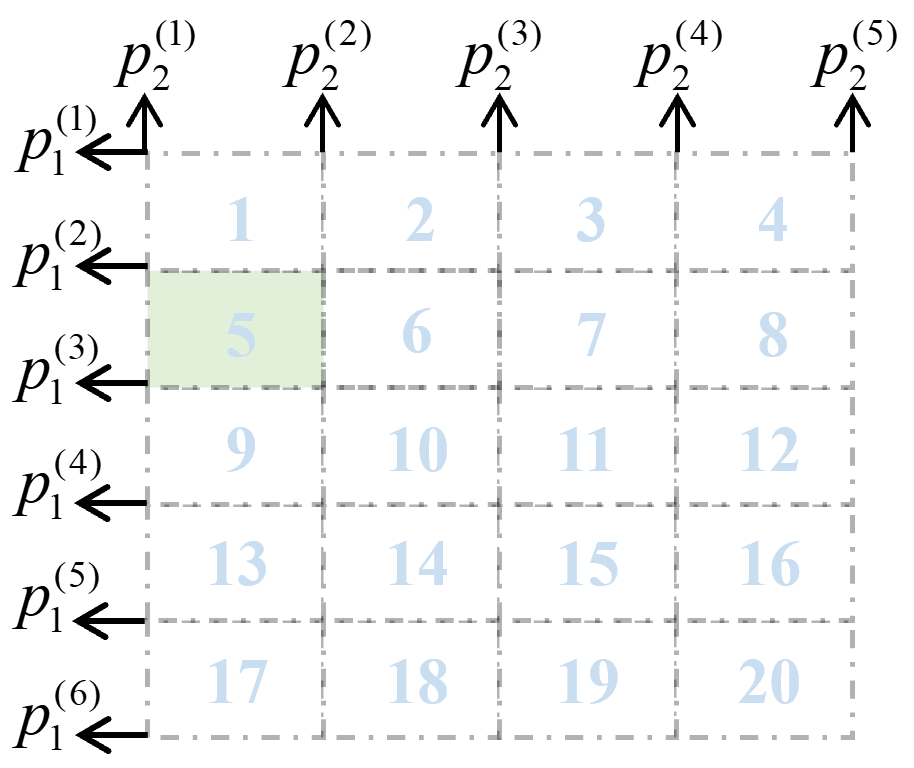}%
\label{fig2_first}}
\hfil
\subfloat[]{\includegraphics[width=1.9in]{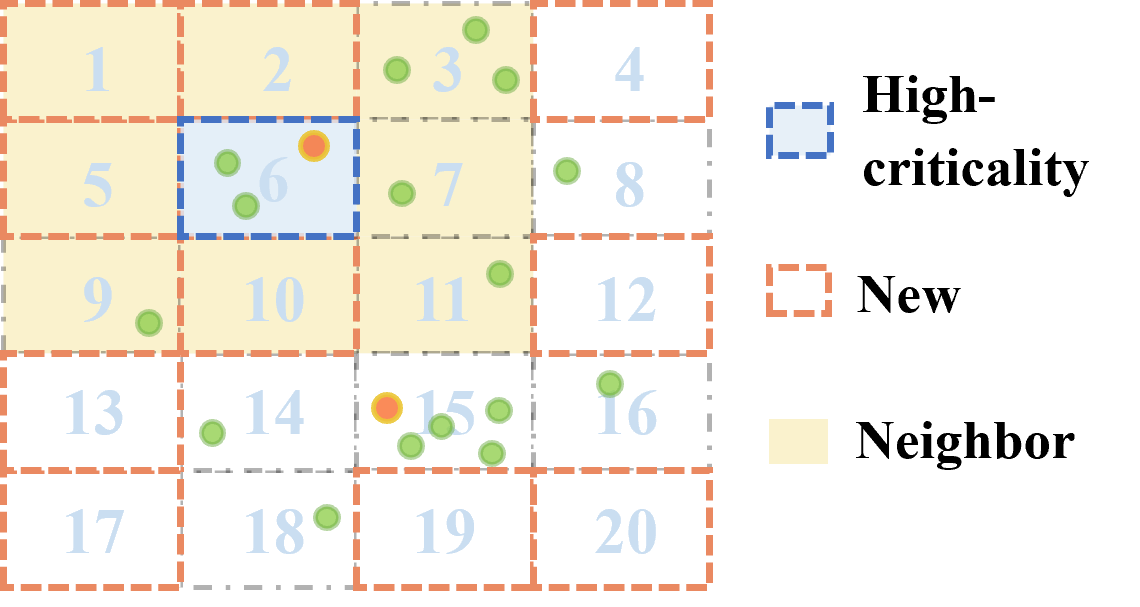}%
\label{fig2_second}}
\caption{Scenario parameter space representation. (a) visualization of space abstraction. (b) visualization of subspace evaluation.}
\label{fig2}
\end{figure}

For clarity, \autoref{fig2}(a) visualizes the space abstraction process of a 2-dimensional scenario parameter space with $m_1=5$ and $m_2=4$. The subspace corresponding to index $\mathbf{k}=(2,1)$ corresponds to $\left[p_1^{(2)}, p_1^{(3)}\right) \times \left[p_2^{(1)}, p_2^{(2)}\right)$. Although this illustrates a 2-dimensional case, the multi-index notation $\mathbf{k} = (k_1,\dots,k_N)$ generalizes to high-dimensional spaces where vector indices become increasingly complex. To simplify visual representation across dimensions, we apply a reversible mapping that converts these multi-indices to sequential integer labels. Thus, subspace $S_{(2,1)}$ is labeled as $S_5$ in \autoref{fig2}(a) and highlighted in green.

For clarity, \autoref{fig2}(a) visualizes the space abstraction of a 2-dimensional scenario parameter space with $m_1=5$ and $m_2=4$. The green-highlighted subspace in this figure corresponds to index $\mathbf{k}=(2,1)$, which defines the coordinate range:
\begin{equation*}
S_{(2,1)} = \left[p_1^{(2)}, p_1^{(3)}\right) \times \left[p_2^{(1)}, p_2^{(2)}\right).
\end{equation*}
Although this demonstrates a 2-dimensional case, the multi-index notation $\mathbf{k} = (k_1,\dots,k_N)$ extends to high-dimensional spaces where vector indices grow increasingly complex. To achieve concise representation of high-dimensional subspace indexing, we apply a reversible mapping that converts these multi-indices to sequential integer labels. Thus, subspace $S_{(2,1)}$ is labeled as $S_5$ in \autoref{fig2}(a).

\subsubsection{Subspace Evaluation Layer}

The space abstraction layer transitions scenario exploration from high-dimensional scenario parameters to lower-dimensional subspaces, significantly reducing exploration complexity. However, the localization efficiency of critical scenarios remains constrained by the lack of quantitative characterization for high-risk subspace distributions and exploration coverage. To overcome this limitation, we introduce a subspace evaluation layer to characterize the criticality and diversity of each subspace. This approach employs three representative metrics:  \emph{spatial criticality} and \emph{neighborhood correlation} jointly guide the localization of critical subspaces, while \emph{test density} guides the localization of diverse subspaces.

\emph{a) Test Density:} Extremes in subspace exploration frequency compromise testing efficiency: over-exploration causes resource waste, while unexplored subspaces pose the risk of missing diverse critical scenarios. To address this, the test density metric is defined to quantify subspace exploration coverage, formally expressed as 
\begin{equation}
D(S_i) = \sum_{j=1}^M \bm{1}_{S_i}(\Gamma(z_j)),
\end{equation}
where $ z_j \in \mathbb{R}^N $ denotes the concrete scenario in the $ j^{th} $ test iteration. $\Gamma$ is the abstraction function that maps scenario $ z_j $ to a subspace within the subspace set $ \mathbb{S} = \{S_1,\ldots,S_K\}$. The indicator function $\bm{1}_{S_i}(\cdot)$ activates (returns 1)  when the abstracted scenario belongs to subspace  $S_i$, otherwise remains 0. The summation index $M \in \mathbb{N}^+$ represents the cumulative count of testing iterations. 

This metric identifies uncovered regions (i.e., diverse subspaces with zero test density) through cumulative testing instances, guiding the exploration to diverse subspaces. For clarity, \autoref{fig2}(b) illustrates the abstraction of a two-dimensional scenario parameter space, where the two dimensions are divided into 5 and 4 intervals, respectively, resulting in 20 subspaces. The test density of subspace $S_6$ is $D(S_6)=3$, indicating three explorations until the $j^{th}$ testing; while  $D(S_{15}) = 5$ reflects more frequent exploration of subspace $S_{15}$.

\emph{b) Spatial Criticality and Neighborhood Correlation:} Neighboring subspaces of critical scenarios within the parameter space often exhibit correlated risk potentials. Consider an autonomous driving example: the subspace $S_i$ represents scenarios with a 0.5-meter lateral distance from the curb, where vehicle trajectories would result in curb contact. Once $S_i$ is identified as a high-criticality subspace, adjacent subspaces such as $S_j$ (e.g., 0.6 meters from the curb) can still generate diverse critical scenarios when combined with steering angle variations. To capture such scenarios, we propose defining two metrics: spatial criticality and neighborhood correlation. These metrics identify core risk subspaces and detect underlying high-criticality subspaces, respectively, thereby guiding exploration directions for scenario generation.

Spatial criticality metric quantifies the risk concentration within subspaces through normalized critical scenario triggering frequency, i.e., 
\begin{equation}
\label{eq:criticality_metric}
\begin{split}
K(S_i) &= \frac{F(S_i)}{D(S_i) }, \\
F(S_i) &= \sum_{j=1}^M \mathbf{1}_{\mathcal{F}}(\Gamma(s_j), S_i),
\end{split}
\end{equation}
where $K(S_i)$ represents the criticality of subspace $S_i$. $F(S_i)$ denotes the number of critical scenarios within subspace $S_i$. $\mathbf{1}_{\mathcal{F}}(\Gamma(s_j), S_i)$ returns 1 if two conditions hold: (i) the abstracted scenario $\Gamma(s_j)$ belongs to $S_i$, and (ii) the scenario triggers critical condition $\mathcal{F}$. For example in \autoref{fig2}(b), $F(S_6) = 1$ indicating the subspace $S_6$ contains one critical scenario, while $F(S_7) = 0$ demonstrates that subspace $S_7$ exhibits no observed critical scenarios. As exemplified in \autoref{fig2}(b), subspace $S_6$ achieves the highest criticality value of 1/3 among all subspaces, marking it as the core risk subspace. Conversely, subspace $S_3$ with zero criticality indicates negligible risk potential, thereby reducing the exploration of this subspace.

The neighborhood correlation metric quantifies spatial dependencies between subspaces in the $N$-dimensional parameter space. For subspace $S_i$, the neighborhood set $\mathcal{N}(S_i)$ consists of adjacent subspaces defined by:
\begin{equation}
\mathcal{N}(S_i) = \left\{ S_j \in \mathbb{S} \mid \bm{c}_j = \bm{c}_i + \bm{\Delta} \text{ for } \bm{\Delta} \in \mathcal{A} \right\},
\label{eq:neighborhood_def}
\end{equation}
where $\mathbb{S}$ denotes the complete set of subspaces, $\bm{c}_i = (d_1,d_2,\dots,d_N)$ represents spatial position of $S_i$, $\bm{c}_j$ represents the coordinate vector of neighboring subspace $S_j$. The dimensional displacement vector $\bm{\Delta}$ is described as
\begin{equation}
    \bm{\Delta} = (\delta_1,\dots,\delta_k,\dots,\delta_N), \bm{\Delta} \in \mathcal{A},
\end{equation}
\begin{equation}
    \mathcal{A} = \{-1,0,+1\}^N \setminus \{\bm{0}\},
\end{equation}
 where each $\delta_k$ corresponds to a unit offset in the $k^{th}$ dimension, and $\mathcal{A}$ defines the set of admissible offset vectors that excludes the zero vector $\bm{0}$ to prevent self-inclusion.

As demonstrated in \autoref{fig2}(b), $S_6$ (identified as the highest-criticality subspace through spatial criticality analysis) has its neighborhood  $N(S_6) = \{ S_1, S_2, S_3, S_5, S_7, S_9, S_{10}, S_{11} \}$. These neighbors represent spatially correlated high-criticality candidates, directing scenario exploration toward these prioritized regions.

\subsection{Dual-space Guided Testing Scenario Generation}

\begin{figure}[!htb]
\centering
\includegraphics[width=8.5cm]{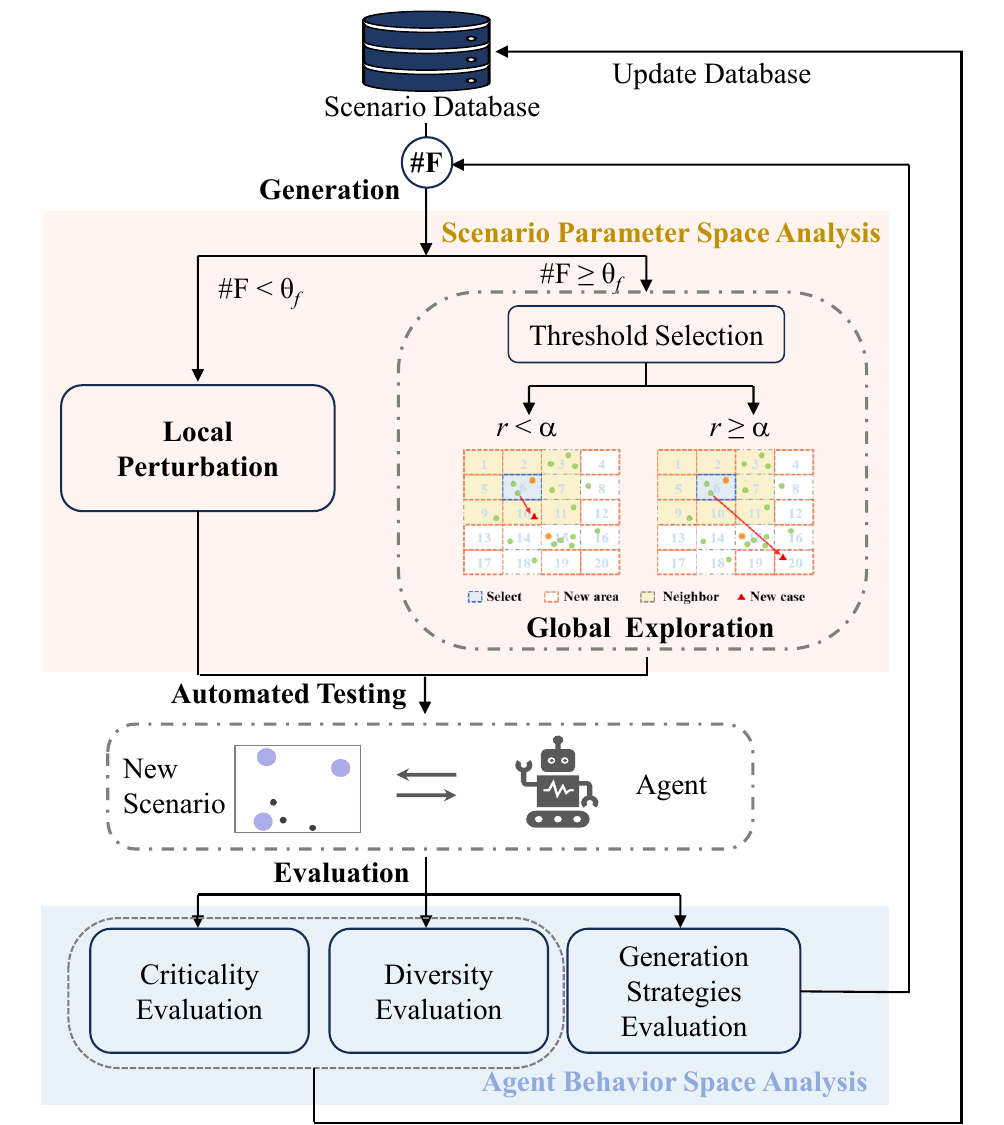}%
\caption{Illustration of dual-space guided generation strategy}
\label{fig:3}
\end{figure}

The hierarchical representation framework of scenario parameter space effectively guides the exploration for diverse and critical scenarios in most cases. However, its exclusive reliance on scenario parameters and criticality labels neglects the dynamic behaviors during decision-making agents' interaction. This oversight may lead to incomplete scenario evaluation. For instance, scenarios within the same parameter subspace may produce drastically different behavioral responses through interactions with decision-making agents. To address this limitation, we propose a dual-space collaborative analysis framework that couples the agent behavior space with the existing parameter space. Following each automated testing execution, we extract decision-making patterns and performance metrics, quantitatively evaluate behavioral criticality and diversity, and utilize the assessment results as feedback to guide the generation of diverse and critical scenarios. The integrated workflow is illustrated in \autoref{fig:3}.

\subsubsection{Scenario Parameter Space Analysis}

The scenario parameter space employs a testing scenario generation strategy combining local perturbation and global exploration to achieve dynamic balance between the quantity and diversity of critical scenarios. Sole reliance on local perturbation leads to redundant exploration of local subspaces, resulting in test resource waste and limitations on global coverage of the parameter space. While pure global exploration improves spatial coverage, it reduces the sufficiency of exploitation in high-potential subspaces, thereby decreasing critical scenario generation efficiency. To address this, a monitoring mechanism based on sliding windows is developed to assess critical scenario generation efficiency. When the estimated generation efficiency exceeds a predefined threshold  ($\#F > \theta$), this implies that the current subspace still contains underdeveloped potential for critical scenarios. Consequently, the local perturbation mode remains active to continue generating testing scenarios. In this mode, base scenarios with high sensitivity from the scenario database are prioritized for random perturbation. The sensitivity metric is defined as Eq.\eqref{eq:sensitivity}, quantifies the behavioral divergence of a scenario caused by minor parameter perturbations. Since the behavior of highly sensitive scenarios exhibits strong responsiveness to parameter variations, minor perturbations can induce significant behavioral pattern mutations, which dominantly trigger critical scenarios through sensitivity-guided exploration compared to undirected random sampling. The sensitivity of scenarios is defined as:
\begin{equation}
\sigma(s_i) = \frac{|\mathcal{T}(s_i) - \mathcal{T}(s_i')|}{\|s_i - s_i'\|},
\label{eq:sensitivity}
\end{equation}
where $\mathcal{T}(s)$ and $\mathcal{T}(s')$ represent the criticality score of scenario $s_i$ and $s_i'$, respectively, and $\|s_i - s_i'\|$ denotes the magnitude of the parameter perturbation.

As the testing process progresses, when the estimated generation efficiency within the sliding window falls below a preset threshold ($\#F < \theta$), this indicates that the current local subspace has reached a saturated exploration state. Persisting with local perturbation mode under such conditions would lead to redundant exploration and reduced testing efficiency.  Therefore, global exploration mode is activated, shifting exploration direction focus to uncovered subspaces in the scenario parameter space. To mitigate resource inefficiency caused by indiscriminate global exploration, a hybrid exploration strategy is developed, employing an adaptive modulation factor  $\alpha \in (0,1)$ to balance directional exploitation and random exploitation. When the random number $r \sim \mathcal{U}(0,1)$ satisfies $r < \alpha$, a directional exploration approach is adopted. Specifically, leveraging information from the hierarchical representation framework, scenarios from Top-K high-criticality subspaces $S_k$ are selected as base scenarios, with exploration concentrated in their neighborhood $\mathcal{N}(S_k) $ and zero-test-density subspaces. Conversely, when $r \geq \alpha$, random exploration is executed across all unexplored subspaces in the entire parameter space to generate testing scenarios. The strategy is formally defined as:
\begin{align}
S_i = \begin{cases}
S_i \in \mathcal{N}(S_k) \cap \{ S \,|\, D(S)=0 \}, & \text{if } r < \alpha \\
S_i \in \{ S \,|\, D(S)=0 \}, & \text{if } r \geq \alpha 
\end{cases}
\end{align}
where $r \sim \mathcal{U}(0,1)$ is a random variable uniformly sampled from $[0,1]$, $ D(\cdot) $ represents the test density of a given subspace, $\mathcal{N}(S_k) $ represents the neighboring subspaces of the sampled high-criticality subspace $S_k$.

\begin{algorithm}
    \SetAlgoLined
    \LinesNumbered
    \DontPrintSemicolon
    \caption{DiCriTest}
    \SetKwInOut{KwIn}{Input}
    \SetKwInOut{KwOut}{Output}
    
   \KwIn{
    Number of scenario database: $N_{scenario}$, Number of tests: $N_{test}$, Environment: $env$\\}
    \KwOut{
    critical testing scenario: $S_{cri}$\\}
    
    \SetKwFunction{Env}{Env}
    \SetKwProg{Fn}{Function}{:}{}
    \Fn{\Env{$env$, $s$}} {
        $\{s'\}, r \leftarrow$ EnvSim($env$, $s$)\\
        \KwRet $\{s'\}$,r
    }
    
    \SetKwFunction{Database} {Database}
    \SetKwProg{Fn}{Function}{:}{}
    \Fn{\Database{$N_\text{scenario}$ }} {
    $s_{init}\gets \emptyset$\\
        \While {Scenario\_number $<$ $N_\text{scenario}$}{
        $\{s\_{init}\}\leftarrow$ Uniform\_Sample($env$)\\
        $\{s_\text{init}', r_\text{init}\} \leftarrow$ {\Env}($s_{init}$)\\
        }
        \KwRet $s_{init}$
    }

    \SetKwFunction{Abstract} {Abstract}
    \Fn{\Abstract{$\mathcal{S}, m$}} {
        \For{$i \in [1,N]$}{
            $\{p_i^{(j)}\} \gets \text{linspace}(a_i, b_i, m+1)$ \;
        }
        $\mathcal{S}_{\mathbf{k}}=\prod_{i=1}^N [p_i^{(k_i)}, p_i^{(k_i+1)}) \quad \forall k_i \in [0,m-1]$\\
         
          \KwRet $\mathcal{S}_{\mathbf{k}}$
     }

    \SetKwFunction{Repr} {Repr}
    \SetKwProg{Fn}{Function}{:}{}
    \Fn{\Repr{{${repr\_dict}$}, $\mathcal{S}_{\mathbf{k}}$, $\text{Critical\_flag}$}} {
        \texttt{${repr\_dict}$}[$\mathcal{S}_{\mathbf{k}}$] $\leftarrow$ ($N$, $D$, $\#F$, $K$)\\
    }
   
    \SetKwFunction{Generation} {Generation}
    \SetKwProg{Fn}{Function}{:}{}
    \Fn{\Generation{$\#F$, $s_{base}$}} {
        \If { $\#F$ $<$ $\theta$}{
        sample $s_i$ from $s_{base}$ with  $K$\\
        generate $s_\text{new}$ from $N_k$ and $D_i$\\
        }
        \Else{
        sample $s_i$ from $s_{base}$ with  $\sigma$\\
        generate $s_\text{new}$ by adding $\triangle s$ to $s_i$
        }
        \KwRet $s_\text{new}$
    }
    
    \SetKwFunction{IsCritical}{IsCritical}
    \SetKwFunction{DiCriTest} {DiCriTest}
    \SetKwProg{Fn}{Function}{:}{}
    \Fn{\DiCriTest{$env$}} {
     ${repr\_dict} \gets \emptyset$ \\
    $s_{base} \leftarrow$ {\Database}($N_\text{scenario}$)\\
    $S_{base} \leftarrow$ {\Abstract}($s_{base}$)\\ 
        \While{test\_num $ < N_{test}$}{
       $s_\text{new}$  $\leftarrow$  {\Generation}(\#F, $s_{base}$)\\
       $S_\text{new}$  $\leftarrow$  {\Abstract}($s_\text{new}$)\\
       ${Critical\_flag}$$\leftarrow$  {\IsCritical}($s_\text{new}$)\\
       \If {${Critical\_flag}$}{
        Add $s_{new}$ to $s_{\text{cri}}$\\
        }
        \ElseIf {$r_\text{new} < r_{i}$ or \text{behavior}($r_{new}$) $<$ $\tau$} {
        Add $s_\text{new}$ to $s_{base}$\\     
        }
        $s_\text{new}$  $\leftarrow$  {\Generation}(\#F, $s_{base}$)\\
       {\Repr}({${repr\_dict}$}, ${s}_{\mathbf{new}}$, ${Critical\_flag}$)\\
       }
       \KwRet $s_{\text{cri}}$
    }

\end{algorithm}
\DecMargin{1em}

\subsubsection{Agent Behavior Space Analysis}

To compensate for prediction inaccuracies in diverse and critical subspaces within the scenario parameter space, a dynamic posteriori evaluation mechanism based on the behavioral data of agents is developed. Once the generated scenarios obtain precise spatiotemporal interaction data through testing, this mechanism drives incremental updates of the database via real-time criticality calculations while guiding autonomous switching between scenario generation modes (local perturbation or global exploration) in subsequent iteration cycles. 

Specifically, by analyzing motion trajectories (position, velocity) and criticality indicators during interactions between the current scenario and the decision-making agent, we calculate its task execution score $\mathcal{T}$, which exhibits a negative correlation with scenario criticality, i.e.,
\begin{equation}
\begin{aligned}
\mathcal{T} = & \sum_{t=0}^{M} \gamma^{t} \left( \alpha f(d_{t}) + \beta g(\Delta v_{t}) \right) - \sum_{k \in \mathcal{F}} \lambda_{f}, \\
\label{eq:behavior score}
\end{aligned}
\end{equation}
where $t$ indexes the interaction step with the decision-making agent, $\gamma^{t} \in (0,1]$ is the discount factor balancing immediate and long-term rewards, $\alpha,\beta \in \mathbb{R}^+$ are weighting coefficients for distance and velocity tracking respectively. The function $f(d_t): \mathbb{R} \to \mathbb{R}$ computes distance-based scores, $g(\Delta v_t): \mathbb{R} \to \mathbb{R}$ evaluates speed tracking accuracy. The penalty term $\sum_{k \in \mathcal{F}} \lambda_f$ aggregates violations across all critical scenarios in $\mathcal{F}$, where $\lambda_f > 0$ is the penalty coefficient for violating the $k^{th}$ critical scenario constraint. 

The task execution scores of newly generated scenarios ($\mathcal{T}_{new}$) and base scenarios ($\mathcal{T}_{base}$) are computed separately. When $\mathcal{T}_{new} < \mathcal{T}_{base}$, this indicates that the current parameter perturbation direction effectively enhances scenario criticality. At this point, the new scenario is incorporated into the scenario database, demonstrating higher potential for evolving into critical scenarios and thereby becoming the optimization benchmark for subsequent iterations.

However, behavioral data from the current single-scenario is insufficient for comprehensive evaluation. To address this, we construct a multi-cycle scenario evaluator that assesses scenarios by synthesizing behavioral data across multiple testing iterations. As suggested in \cite{pang2022mdpfuzz}, we employ Gaussian Mixture Model (GMM) to probabilistically characterize both the newly generated trajectory data and historical trajectories from all prior testing cycles: 
\begin{equation}
    \Pr\left( \{ T_t \}_{t=0}^M \right) = 
    \mathrm{GMM}^s(T_0) 
    \prod_{t=0}^{M-1} 
    \frac{
        \mathrm{GMM}^c \left( T_{t+1} \parallel T_t \right)
    }{
        \mathrm{GMM}^s(T_t)
    },
    \label{eq:trajectory_probability}
\end{equation}
where $\mathrm{GMM}^s(T_t) = \sum_{k=1}^K \phi_k^{(s)} {N}(T_t | \mu_k^{(s)}, \Sigma_k^{(s)})$ models single-state distributions with mixture weights $\phi_k^{(s)} \in (0,1)$, and $\mathrm{GMM}^c$ captures state transitions through joint distributions $\sum_{k=1}^K \phi_k^{(c)} \mathcal{N}([T_t, T_{t+1}] | \mu_k^{(c)}, \Sigma_k^{(c)})$. The model parameters $\{\phi_k^{(s)}, \mu_k^{(s)}, \Sigma_k^{(s)}, \phi_k^{(c)}, \mu_k^{(c)}, \Sigma_k^{(c)}\}$ are iteratively updated via an online expectation-maximization (EM) algorithm.

The trajectory probability density value $\Pr(\{T_t\})$ quantifies the distributional divergence between current scenarios and the historical scenarios, where lower values indicate higher behavioral novelty. This inverse correlation emerges because frequently observed scenarios predominantly occupy high-density regions of the GMM. To mitigate redundant scenario accumulation, scenarios satisfying $\Pr(\{T_t\}) < \lambda$ are deemed novel and incorporated into the database, where the threshold $\lambda$ is dynamically adapted based on historical scenario distributions. This mechanism ensures continuous enrichment of scenario diversity while preventing over-saturation of similar patterns.

Furthermore, we propose an adaptive sliding window mechanism for online evaluation of scenario generation strategies. This approach quantifies strategy effectiveness by monitoring the critical scenario generation rate within dynamically updated observation intervals, thereby addressing temporal distribution shifts through adaptive window resizing. Due to time-varying distribution shifts in long-term historical data and increased online computational complexity, we evaluate generation efficiency via the metric $\#F$:
\begin{equation}
\#F = \frac{N_{\text{cri}}}{W},
\end{equation}
where $W$ is the window size and $N_{cri}$ counts critical scenarios. The adaptive window directly governs mode switching through a dynamically adjusted threshold $\theta$: when $\#F < \theta$, the global exploration generation mode is activated, while local perturbation mode is selected when $\#F \geq \theta$.

\subsection{Diversity Evaluation Metrics}

In the generation of critical scenarios for decision-making agents, the diversity of scenarios proves equally crucial as their quantity for comprehensive validation. Solely focusing on accumulating scenario quantities may lead to testing resources being dominated by a large number of parameter-similar redundant scenarios. While solely relying on diversity metrics in the agent behavior space can capture diversity in behavioral patterns, they fail to quantify the comprehensive exploration of high-dimensional parameter spaces, causing the generation process to become trapped in local regions. Therefore, we propose parameter-behavior co-driven diversity metrics for evaluating critical scenarios, including coverage, initial scenario distance, and trajectory similarity. In the scenario parameter space, to quantify the coverage completeness of the exploration process, we propose a coverage metric $cvg$ based on the hierarchical representation framework, which counts the number of subspaces with non-zero testing density, i.e.,
\begin{equation}
\label{eq:coverage}
cvg = \sum_{i=1}^{\mathbf{K}} \bm{1}(D(S_i) > 0),
\end{equation}
where $\mathbf{K}$ denotes the total number of subspaces, $D(S_i)$ represents the testing density of subspace $S_i$ , and $\bm{1}(\cdot)$ is the indicator function that returns 1 if the condition $D(S_i) > 0$ is satisfied and 0 otherwise.

To quantify the aggregation degree of scenario distributions, we introduce an initial scenario distance metric $dis$ based on the mean Euclidean distance between scenarios, i.e.,
\begin{equation}
\label{eq:initial_distance}
dis = \frac{2}{Q} \sum_{i=1}^{Q-1} \sum_{j=i+1}^{Q} \| {s}_i - {s}_j \|_2,
\end{equation}
where $Q$ is the total number of critical scenarios, ${s}_i \in \mathbb{R}^N$ represents the $N$-dimensional parameter vector of the $i$-th critical scenario, and $\|\cdot\|_2$ denotes the Euclidean distance operator. 

In the behavioral space, a trajectory similarity metric defined by Eq.\eqref{eq:trajectory_probability} evaluates scenario variations by calculating the mean value across all generated scenarios, where higher results signify concentrated behavioral distributions and consequently lower diversity.

\begin{table}[t]
\caption{Parameter Settings of DiCriTest}
\label{tab:parameters}
\centering
\scriptsize
\setlength{\tabcolsep}{6pt}
\renewcommand{\arraystretch}{1.1}

\begin{tabular}{@{}ccc c c c@{}}
\toprule
\multicolumn{1}{c}{\textbf{Environment}} & 
\multicolumn{1}{c}{\textbf{Model}} & 
\multicolumn{1}{c}{\textbf{\makecell{Scenario\\DataBase}}} & 
\multicolumn{1}{c}{\textbf{\makecell{Test\\Number}}} & 
\multicolumn{1}{c}{\textbf{\makecell{Window\\Size}}} & 
\multicolumn{1}{c}{$\bm{\theta}$} \\
\midrule
ACAS\_Xu    & DNN  & 2000 & 100000& 1000 & 0.001\\
CoopNavi    & MARL & 1000 & 10000& 1000 & 0.13\\
BipedalWalker & RL  & 1000 & 1000& 100  & 0.14\\
RLCARLA     & RL   & 100  & 600  & 100  & 0.1\\
ILCARLA     & IL   & 100  & 600  & 100  & 0.2\\
\bottomrule
\end{tabular}
\end{table}

\section{\textbf{Experiments} }

In this section, we conducted a series of experiments to validate the performance of our proposed dual-space guided testing framework, DiCriTest. The key findings are as follows: (i) Under the same number of iterations, our method generated 56.23\% more critical scenarios on average than baseline methods, outperforming them in efficient critical scenario generation; (ii) Through joint diversity analysis of the scenario parameter space and agent behavior space, our method demonstrated enhanced diversity in generated critical scenarios; (iii) Visualization analysis on critical scenarios' spatial distribution further confirmed that our method generates more critical scenarios while maintaining broad coverage; (iv) Comparative experiments further verified the effectiveness of hybrid scenario generation strategy.

\subsection{Environment }
To comprehensively evaluate the performance of the proposed DiCriTest, we select four widely used decision-making environment in this paper. The details of each environment are described as follows.

ACAS\_Xu\cite{marston2015acas}: This environment equips drones and small aircraft with deep neural networks (DNNs) to generate real-time collision avoidance instructions. By analyzing flight states, the aircraft dynamically selects evasive actions that prevent collisions while prioritizing mission continuity.

CoopNavi\cite{lowe2017multi}:  This environment trains collaborative multi-agent groups to navigate shared spaces. Agents jointly optimize path planning using reinforcement learning (RL), avoiding collisions with obstacles and each other, while ensuring all participants reach their targets efficiently.

BipedalWalker\cite{kuznetsov2020controlling}: Designed in OpenAI Gym, this environment challenges a bipedal robot to learn stable locomotion. Through RL, the robot develops energy-efficient gaits that balance walking stability, task completion, and motion smoothness.

RLCARLA and ILCARLA\cite{dosovitskiy2017carla}: Built on the CARLA simulator, these environments test autonomous vehicles in urban scenarios. Agents trained via RL or imitation learning (IL) must navigate dynamic traffic, comply with rules, avoid obstacles, and optimize routes under real-world constraints.

\begin{figure}[!htb]
\centering
\includegraphics[width=8.5cm]{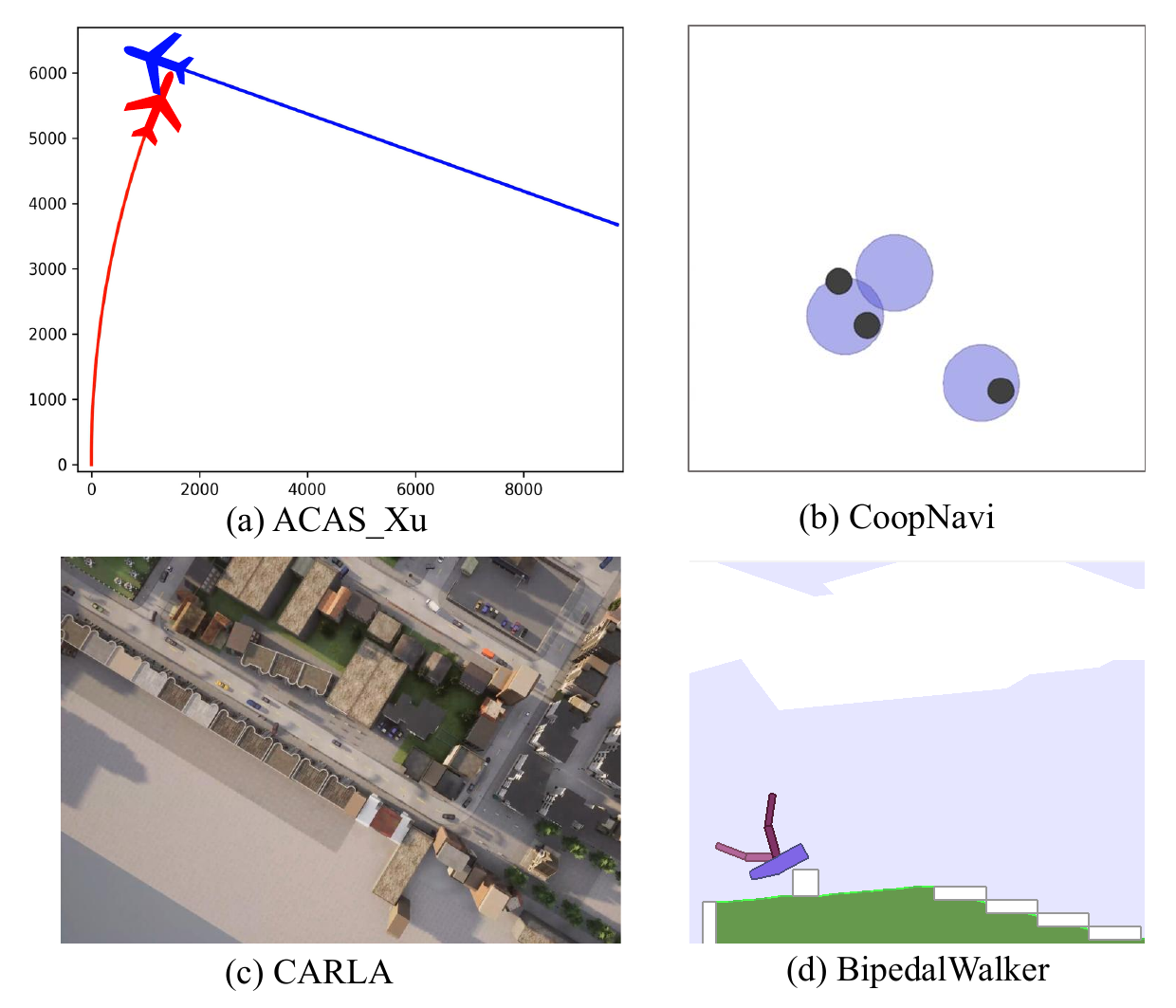}%
\caption{Critical scenarios in four environments}
\label{fig_env}
\end{figure}

The critical scenario diagrams generated by the interactions between the four representative environments and their decision-making agents are illustrated \autoref{fig_env}. The parameter settings for the DiCriTest algorithm are provided in the \autoref{tab:parameters}.

Baseline Methods:  (i) Random testing generates scenarios by uniform sampling of the scenario parameter space and is used as a baseline method in the experiments. However, in the CARLA simulation platform, vehicle positioning is constrained by high-precision maps, requiring testing scenarios to be selected from a predefined library. As a result, random testing is not included in the comparison analysis of CARLA. (ii) MDPFuzz is recognized as the SOTA general-purpose black-box fuzz testing framework, specifically designed for testing decision systems. In this letter, we adaptively modified its open-source \cite{pang2022mdpfuzz} implementation to satisfy our experimental requirements.

\subsection{Evaluation of Effectiveness and Diversity }

\captionsetup[subfloat]{labelformat=empty,skip=0pt}

\begin{figure*}[!t]
\centering

\subfloat{\includegraphics[width=0.198\textwidth,trim=3pt 3pt 3pt 3pt,clip]{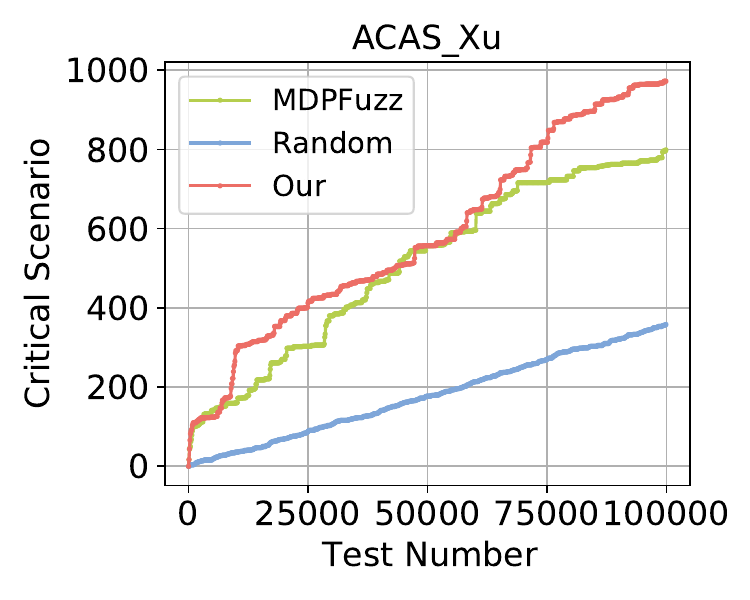}}\hspace{-0.15em}%
\subfloat{\includegraphics[width=0.198\textwidth,trim=3pt 3pt 3pt 3pt,clip]{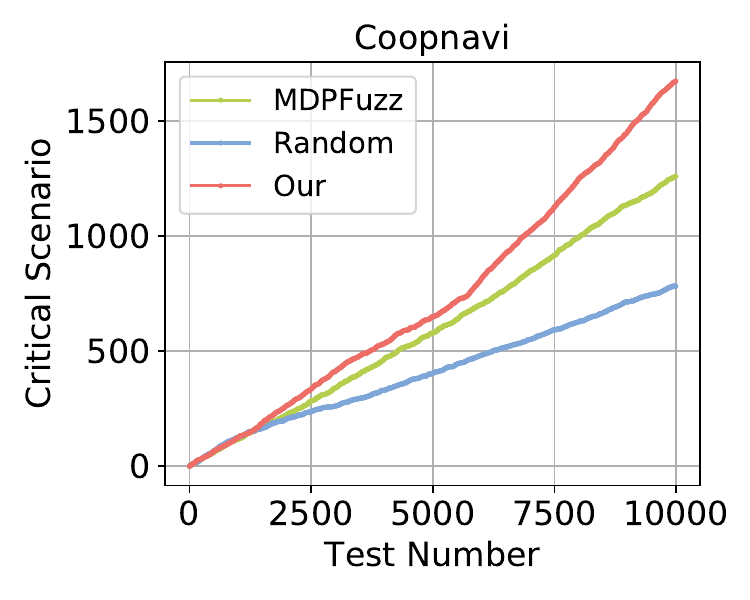}}\hspace{-0.15em}%
\subfloat{\includegraphics[width=0.198\textwidth,trim=3pt 3pt 3pt 3pt,clip]{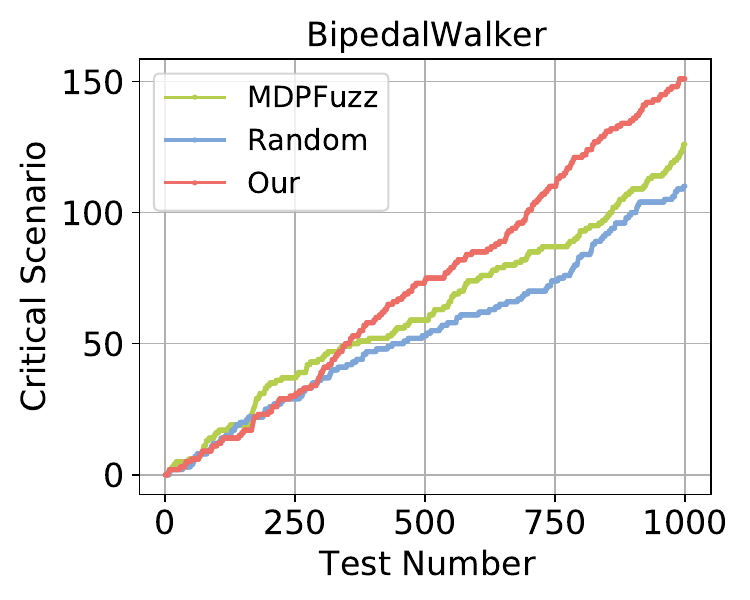}}\hspace{-0.15em}%
\subfloat{\includegraphics[width=0.198\textwidth,trim=3pt 3pt 3pt 3pt,clip]{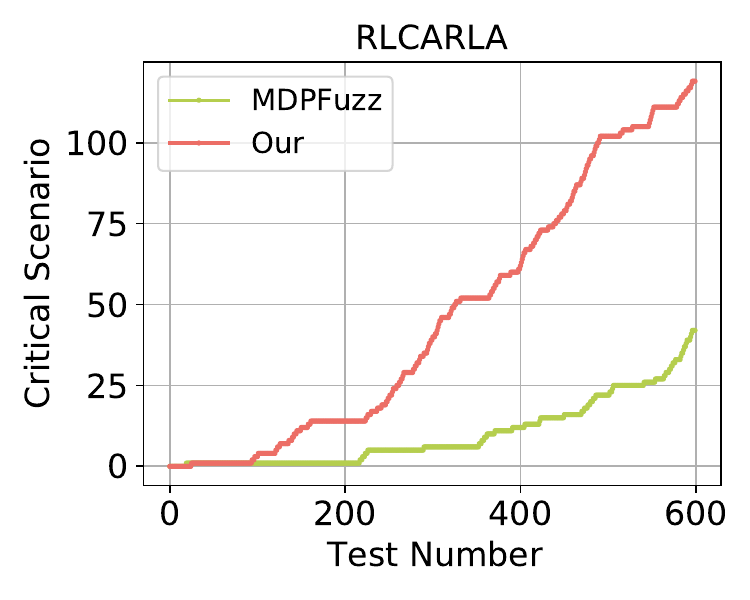}}\hspace{-0.15em}%
\subfloat{\includegraphics[width=0.198\textwidth,trim=3pt 3pt 3pt 3pt,clip]{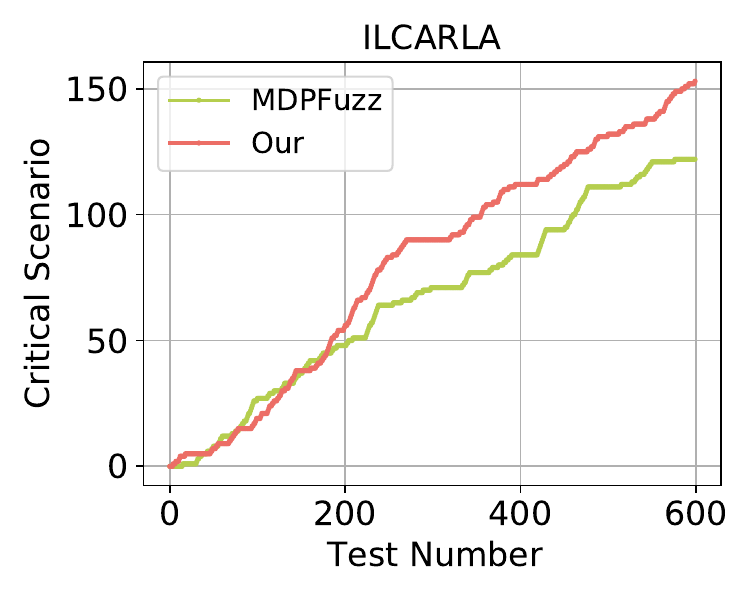}}

\vspace{-5pt}
\caption{Comparison of test number and critical scenario between our method and baselines}
\label{fig:critical}
\end{figure*}

\captionsetup[subfloat]{labelformat=empty,skip=0pt}

\begin{figure*}[!t]
\centering

\subfloat{\includegraphics[width=0.198\textwidth,trim=3pt 3pt 3pt 3pt,clip]{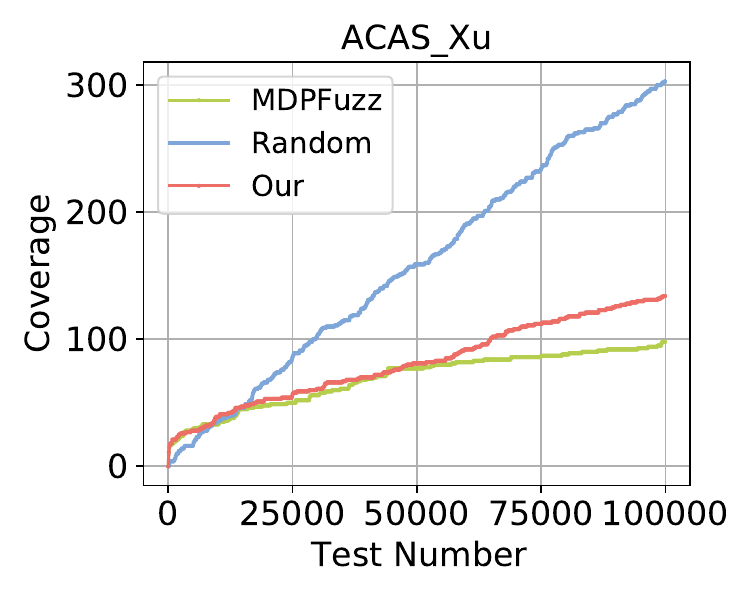}}\hspace{-0.15em}%
\subfloat{\includegraphics[width=0.198\textwidth,trim=3pt 3pt 3pt 3pt,clip]{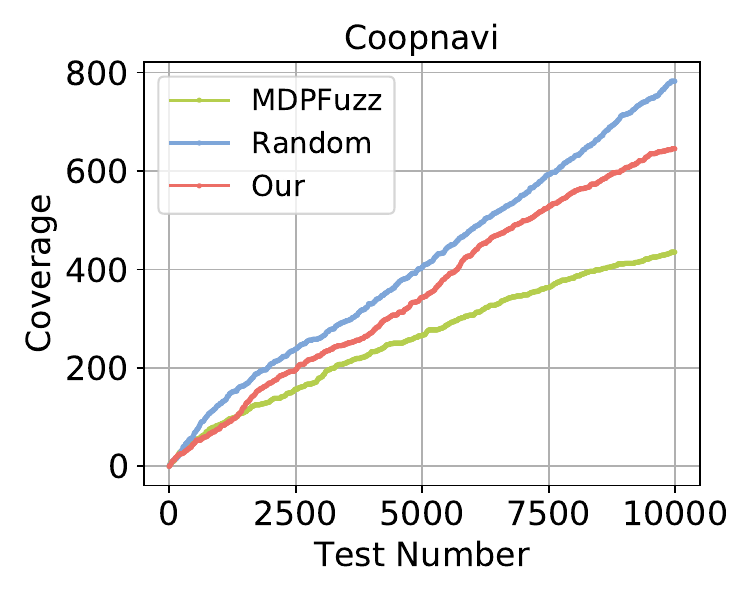}}\hspace{-0.15em}%
\subfloat{\includegraphics[width=0.198\textwidth,trim=3pt 3pt 3pt 3pt,clip]{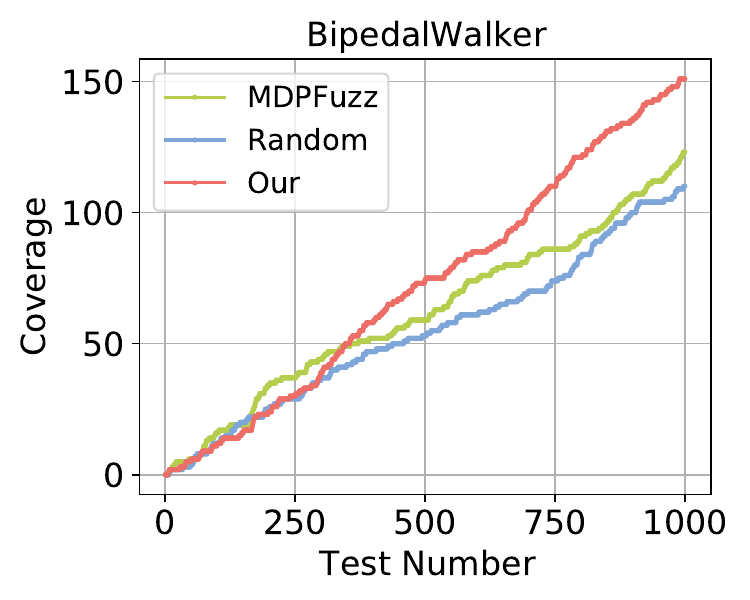}}\hspace{-0.15em}%
\subfloat{\includegraphics[width=0.198\textwidth,trim=3pt 3pt 3pt 3pt,clip]{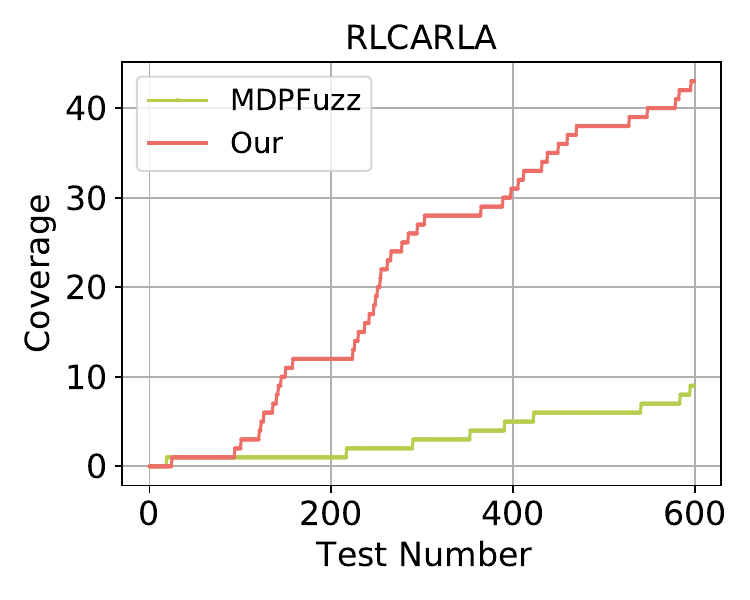}}\hspace{-0.15em}%
\subfloat{\includegraphics[width=0.198\textwidth,trim=3pt 3pt 3pt 3pt,clip]{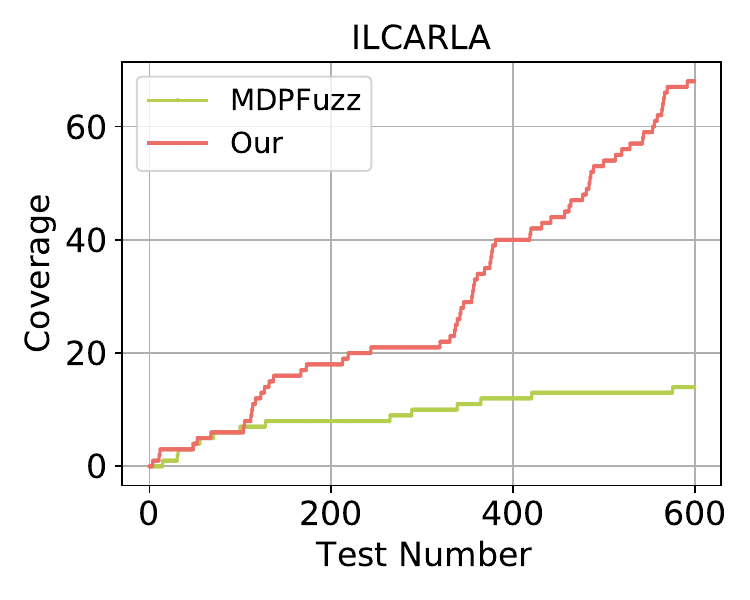}}

\vspace{-5pt}
\caption{Comparison of test number and scenario parameter space coverage between our method and baselines}
\label{fig:cvg}
\end{figure*}

This section conducts a systematic evaluation of critical scenario generation performance between our DiCriTest framework and baseline methods under the same testing iterations, quantifying both critical scenarios generation capacity and diversity of critical scenarios. Distribution characteristics are further analyzed through dimensionality-reduced visual embeddings of the scenario parameter space, providing geometric evidence of coverage completeness.

First, we compare the critical scenario generation capabilities of three testing methods under the same number of testing iterations. As shown in \autoref{fig:critical}, DiCriTest demonstrates significant advantages over baseline methods across five representative testing environments. Specifically, compared to the random testing method, DiCriTest achieves critical scenario generation rate improvements of 171.79\%, 113.67\%, and 37.27\% in the ACAS\_Xu, CoopNavi, and BipedalWalker environments, respectively. In contrast, when compared to MDPFuzz, DiCriTest delivers performance improvements of 21.78\%, 30.81\%, 19.84\%, 183.33\%, and 25.41\% in the ACAS\_Xu, CoopNavi, BipedalWalker, RLCARLA, and ILCARLA environments, respectively. The experimental results indicate that the random method, due to its uniform exploration in the scenario parameter space without criticality guidance, generates the fewest effective critical scenarios. Although MDPFuzz quantifies criticality based on sensitivity and selects high-criticality base scenarios, its single-mode exploration easily traps the parameter space in local optima, resulting in excessive resource consumption in inefficient regions. In contrast, DiCriTest achieves efficient exploration through a dual-space collaborative mechanism: it dynamically locates unexplored high-criticality subspaces based on parameter space distributions, while adaptively selecting global exploration or local perturbation strategies via behavioral space feedback. This enables directional migration of testing resources from saturated regions to unexplored high-risk subspaces, thereby maximizing critical scenario generation efficiency within limited iterations.

Although the differences in the number of critical scenarios generated by various methods have been verified, solely pursuing quantity may lead to scenario redundancy, as repeated scenarios will significantly reduce testing efficiency. To evaluate the diversity of the generated critical scenarios, we further introduce parameter-behavior co-driven diversity metrics (coverage, initial scenario distance, trajectory similarity) for joint quantification. 

\autoref{tab:results} presents the results of the parameter-behavior co-driven diversity metrics, showing that DiCriTest demonstrates significant advantages over MDPFuzz in most cases across five testing environments. Although the random method achieves superior diversity performance through uniform exploration, it generates the fewest critical scenarios among all compared methods, as shown in \autoref{fig:critical}. This demonstrates the fundamental limitation of unguided random search in effectively balancing the diversity-generation efficiency trade-off. Therefore, the following analysis will focus on comparing DiCriTest and MDPFuzz based on the dual dimensions of the parameter space (coverage, initial scenario distance) and the trajectory similarity metric in the behavior space, emphasizing their dual-space collaborative mechanisms.

The parameter space coverage metric is used to measure the distribution breadth of critical scenarios in the parameter space. Experimental data show that DiCriTest achieves significant improvements over MDPFuzz across all five testing environments, with its critical scenario coverage metrics increasing by 37.76\%, 36.36\%, 22.76\%, 366.67\%, and 378.57\%, respectively, as shown in \autoref{fig:cvg}. This advantage stems from the two mode collaborative mechanism: the global exploration mode proactively identifies and covers diverse high-risk subspaces through a hierarchical representation framework, immediately exploring diverse subspaces once the current detection region becomes saturated, while the local perturbation mode enables refined exploration within known high-risk regions.

The initial scenario distance metric reflects the discrete degree of critical scenario in the parameter space. DiCriTest improves this metric by 1.94\% in CoopNavi, 35.31\% in RLCARLA, and 13\% in ILCARLA, proving that its cross-region exploration strategy effectively disperses the distribution of critical scenarios. Notably, in BipedalWalker, the initial scenarios are represented by discrete terrain numbers, which lack interpretable spatial relationships for distance calculation. Thus, this metric was excluded from evaluation. In the ACAS\_Xu environment, although the initial scenario distance metric slightly decreases by 11.36\%, the substantial 37.76\% improvement in coverage suggests that the global exploration mode prioritizes continuous coverage of adjacent high-risk subspaces. While this strategy leads to localized distribution concentration, it significantly expands the coverage boundaries of high-risk scenarios.

The behavioral space trajectory similarity metric quantifies the differences in agent behavior patterns triggered by critical scenarios. DiCriTest reduces this metric by 28.6\% in ACAS\_Xu, 19.3\% in CoopNavi, and 34.7\% in ILCARLA, demonstrating significantly superior behavioral space diversity compared to MDPFuzz. This improvement benefits from a closed-loop mechanism integrating prior exploration guidance and posteriori feedback correction, which dynamically computes behavioral patterns and blocks scenarios with similar behavioral patterns from being stored, thereby preventing redundant exploration. In BipedalWalker and RLCARLA, the trajectory similarity is slightly higher compared to MDPFuzz. However, when combined with the significant improvements in parameter space coverage and initial scenario distance metrics, this behavioral similarity is attributed to the discovery of analogous failure-inducing modes across distinct parameter subspaces, rather than inherent flaws in the methodology. Specifically, the observed convergence in behavioral patterns arises from environmental constraints that dominate agent responses in these subspaces. In contrast, MDPFuzz’s exploration strategy fails to systematically identify such cross-subspace failure modes, further validating DiCriTest’s superiority in balancing diversity and coverage.

In summary, a comprehensive quantitative evaluation has been conducted on both the critical scenario generation capability and diversity. The experimental results demonstrate that our dual-space guided framework significantly enhances critical scenario generation capability. Furthermore, the integrated analysis of parameter-behavior co-driven diversity metrics confirms our superiority in diversity of critical scenarios.

\begin{table}[!htb]
\caption{Diversity of Critical Scenarios}
\label{tab:results}
\centering
\footnotesize
\setlength{\tabcolsep}{3.2pt}

\begin{tabular}{@{}c c cc c }
\toprule
\multirow{2}{*}{\textbf{Environment}} & 
\multirow{2}{*}{\textbf{Method}} & 
\multicolumn{3}{c}{\textbf{Metrics}} \\   
\cmidrule(r{2pt}){3-5}
& & \textbf{Coverage}& \textbf{Distance}& \textbf{Trajectory}\\
\midrule

ACAS\_Xu 
& MDPFuzz & 98 & 7007 & $5.50\times10^{-3}$
\\
& Random  & 303 & 7045 & $1.44\times10^{-7}$
\\
& Our     & 135 & 6211 & $4.05\times10^{-3}$
\\
\midrule
CoopNavi 
& MDPFuzz & 473 & 1.03 & $1.07\times10^{-2}$
\\
& Random  & 783& 1.05& $4.87\times10^{-5}$\\
& Our     & 645 & 1.05 & $5.87\times10^{-5}$
\\
\midrule

BipedalWalker 
& MDPFuzz & 123 & -- & $1.90\times10^{-1}$
\\
& Random  & 110& -- & $1.50\times10^{-1}$\\
& Our     & 151 & -- & $2.51\times10^{-1}$
\\
\midrule

RLCARLA 
& MDPFuzz & 9 & 119.8 & $2.76\times10^{-9}$
\\
& Our     & 42 & 162.1 & $1.13\times10^{-5}$
\\
\midrule

ILCARLA 
& MDPFuzz & 14 & 144.6 & $2.34\times10^{-9}$
\\
& Our     & 67 & 163.4 & $6.94\times10^{-11}$\\
\bottomrule
\end{tabular}
\end{table}

\captionsetup[subfloat]{labelformat=empty,skip=0pt}

\begin{figure*}[!t]
\centering

\subfloat{\includegraphics[width=0.198\textwidth,trim=3pt 3pt 3pt 3pt,clip]{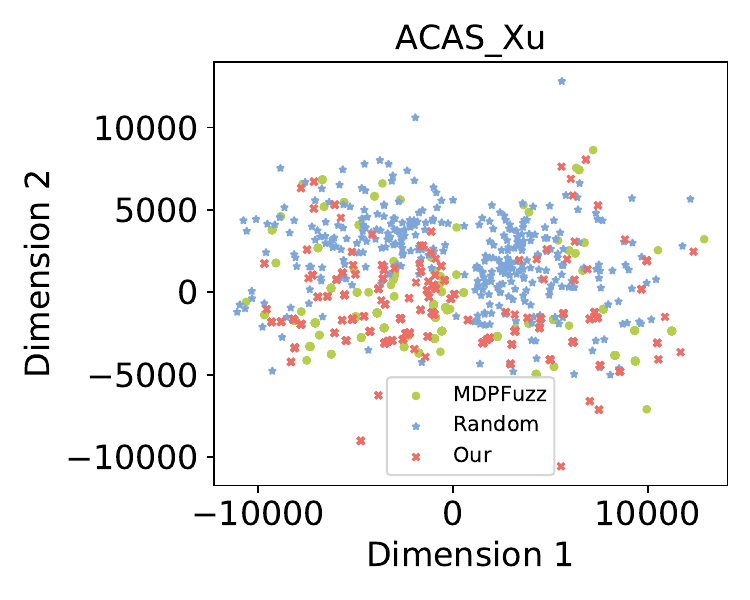}}\hspace{-0.05em}%
\subfloat{\includegraphics[width=0.198\textwidth,trim=3pt 3pt 3pt 3pt,clip]{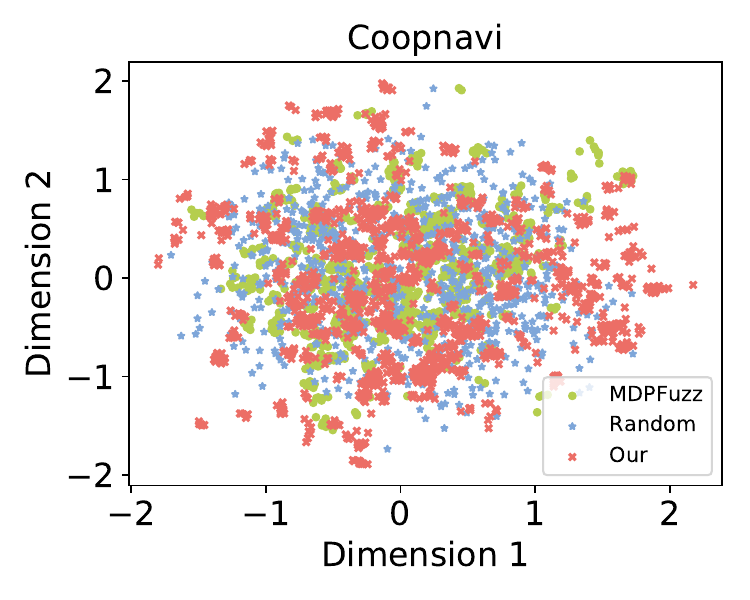}}\hspace{-0.05em}%
\subfloat{\includegraphics[width=0.198\textwidth,trim=3pt 3pt 3pt 3pt,clip]{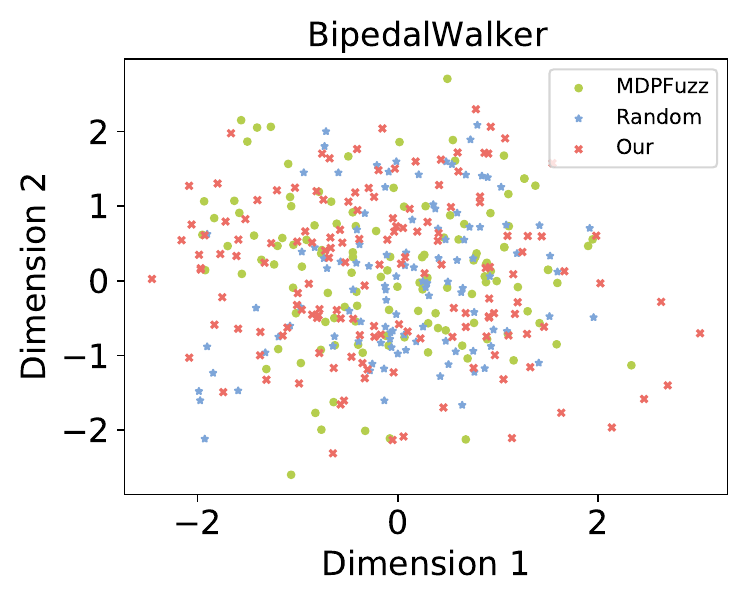}}\hspace{-0.05em}%
\subfloat{\includegraphics[width=0.198\textwidth,trim=3pt 3pt 3pt 3pt,clip]{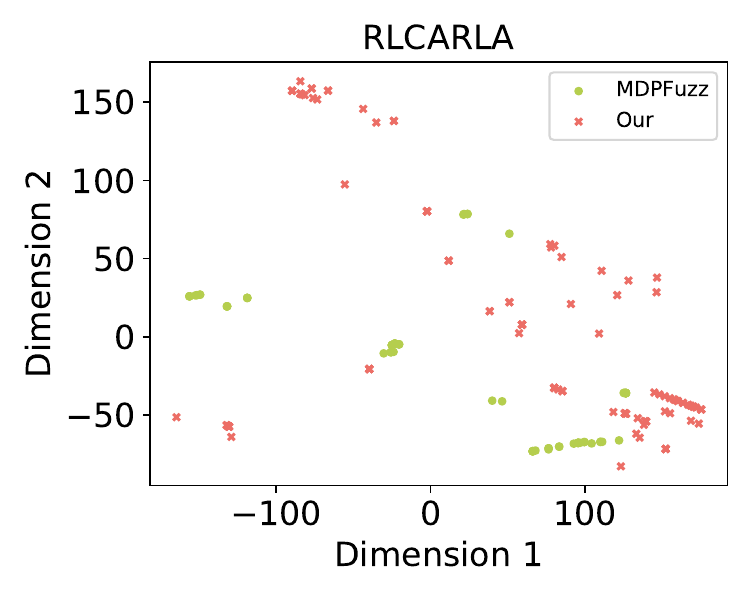}}\hspace{-0.05em}%
\subfloat{\includegraphics[width=0.198\textwidth,trim=3pt 3pt 3pt 3pt,clip]{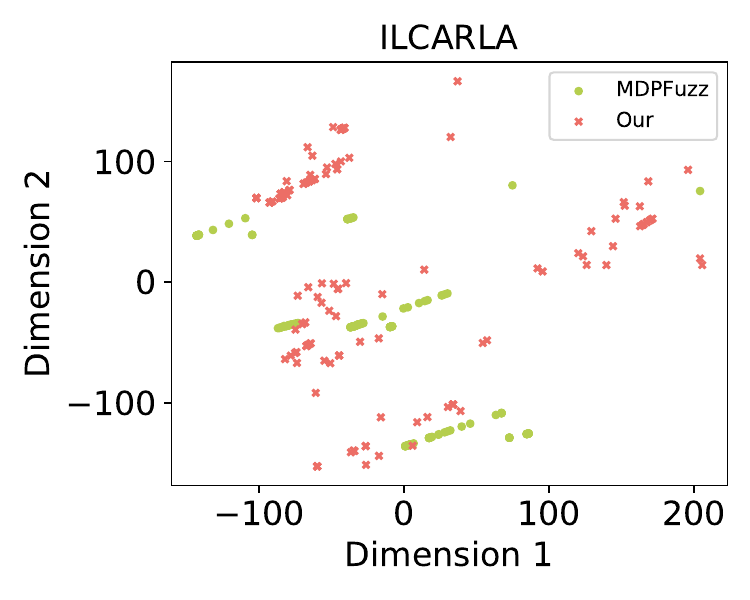}}

\vspace{-5pt}
\caption{Comparison of scenario spatial distribution between our method and baselines}
\label{fig:spatial distribution}
\end{figure*}

To spatially interpret these quantitative advantages and uncover their underlying mechanisms, we visualize the critical scenario distributions via dimensionality reduction, as illustrated in \autoref{fig:spatial distribution}. Compared to baseline methods, two distinct characteristics of the critical scenarios generated by DiCriTest can be observed. First, the critical scenarios exhibit significantly broader spatial coverage than baselines, probing peripheral high-risk subspaces that conventional methods fail to explore due to high-dimensional complexities. Second, the distribution of critical scenarios displays clustered band-like features, exposing latent defects in decision-making agents along stability boundaries. These spatial characteristics further validate the superiority of the dual-space guided testing scenario generation. Real-time integration of the hierarchical representation framework-based global exploration mode enables proactive localization of high-criticality scenarios awaiting exploration, achieving comprehensive coverage. Simultaneously, the local perturbation mode enables thorough exploration in high-criticality regions, which forms clustered band-like distributions. These two modes synergistically ensure that scenario generation achieves both broad coverage of edge regions and enables thorough exploration along stability boundaries, thereby achieving a breakthrough in spatial coverage breadth and detection precision.

\subsection{Effectiveness of Hybrid Scenario Generation Strategy }

This section systematically evaluates hybrid scenario generation strategy across five decision-making benchmarks: ACAS\_Xu, CoopNavi, BipedalWalker, RLCARLA, and ILCARLA. Focusing on the ACAS\_Xu, experiments are conducted with $\alpha \in [0, 0.2, 0.4, 0.6, 0.8, 1]$ over 100,000 testing iterations. As shown in \autoref{tab:hybrid_performance}, the generated testing scenarios are quantitatively analyzed through four metrics: criticality, coverage, initial scenario distance and trajectory similarity. 

To balance the quantity and diversity of critical scenarios, we propose the scoring function:
\begin{equation}
\label{eq:score}
\text{score} = \omega_1\cdot cri + \omega_2\cdot cvg + \omega_3 \cdot dis + \omega_4 \cdot \frac{1}{traj}
\end{equation}
where $cri$, $cvg$, $dis$, and $traj$ denote the criticality , coverage, initial scenario distance, and trajectory similarity, respectively, with weighting coefficients $\omega_1=0.5$, $\omega_2=0.2$, $\omega_3=0.1$, and $\omega_4=0.2$.

Experimental results demonstrate that the adaptive regulation factor $\alpha=0.8$ achieves optimal trade-off between quantity and diversity of critical scenarios, as quantified in \autoref{tab:hybrid_performance}. The hybrid scenario generation strategy governs the quantity-diversity balance of critical scenarios through parameter $\alpha$ as follows:  When $\alpha \rightarrow 1$, the criticality-driven exploitation strategy intensively explores high-risk neighbor subspaces, generating more critical scenarios, but reduces spatial diversity due to localized clustering effects; conversely, when $\alpha \rightarrow 0$, diversity-oriented exploration prioritizes under-sampled subspaces, achieving higher initial scenario distance metrics but suffering a decline in critical scenario quantity from unconstrained criticality sampling. This hybrid scenario generation enables adaptive regulation of exploration strategies between focused critical subspaces and global parameter space coverage, achieving high-efficiency and full-coverage testing scenario generation in decision-making agent verification.

\begin{table}[!htb]
\caption{Performance of Hybrid Strategy under Various $\alpha$ Values}
\label{tab:hybrid_performance}
\centering

\begin{tabular}{@{} c c c c c c @{}}
\toprule
\multicolumn{1}{c}{$\alpha$} & 
\multicolumn{1}{c}{Critical Scenarios} & 
\multicolumn{1}{c}{Coverage} & 
\multicolumn{1}{c}{Distance} & 
\multicolumn{1}{c}{\makecell{Trajectory\\ ($\times10^{-3}$)}} & 
\multicolumn{1}{c}{Score} \\ 
\midrule
0       & 752& 98& 6775.98& 0.66& 0.46\\
0.2     & 585& 113& 7343.27& 8.74& 0.14\\
0.4& 636& 111& 7529.82& 1.41& 0.30\\
0.6     & 718& 130& 6702.47& 1.94& 0.39\\
0.8& 973& 135& 6211.37& 4.05& \textbf{0.66}\\
1       & 820& 150& 6369.71& 1.01& 0.64\\
\bottomrule
\end{tabular}
\vspace{-2mm}
\end{table}

\section{Conclusion}

In the testing of decision-making agents, the precise localization of sparse critical scenarios within the high-dimensional scenario parameter spaces remains a core bottleneck constraining testing efficiency. To address this challenge, this work proposes a dual-space guidance framework for generating diverse critical scenarios, which integrates hierarchical representation of scenario parameter space with dynamic feedback from agent behavior space to efficiently uncover vulnerabilities in decision-making agents. A hierarchical representation framework is constructed in the scenario parameter space to predict subspaces exhibiting both diversity and criticality, guiding exploration directions for hybrid generation strategies combining localized perturbations and global exploration. Furthermore, evaluations in the behavior space adaptively correct subspace prediction biases, enabling incremental iterative optimization of the scenario database and adaptive selection of generation strategies. Extensive experimental results demonstrate that the proposed method outperforms baseline approaches in both the quantity and diversity of generated critical scenarios, providing a promising solution for efficient and comprehensive testing of decision-making agents.


\end{document}